\documentclass[10pt,twocolumn,letterpaper]{article}
\usepackage{dsfont}
\usepackage{multirow}
\usepackage{graphicx} 
\usepackage{booktabs} 
\usepackage{float}  
\usepackage{dblfloatfix}
\usepackage[table]{xcolor}
\usepackage{arydshln}

\usepackage{cvpr}              

\definecolor{cvprblue}{rgb}{0.21,0.49,0.74}
\usepackage[pagebackref,breaklinks,colorlinks,allcolors=cvprblue]{hyperref}

\def\paperID{22478} 
\def\confName{CVPR}
\def\confYear{2026}

\title{CacheFlow: Compressive Streaming Memory for Efficient Long-Form Video Understanding}

\author{
Shrenik Patel\thanks{Equal contribution. Correspondence to shrenik.d.patel@rutgers.edu, daivik.d.patel@rutgers.edu} \\ 
\ Rutgers University \\
\and
Daivik Patel\footnotemark[1] \\
\ Rutgers University \\ 
}

\begin{document}
\maketitle

\begin{figure*}[t]
  \centering
  \includegraphics[width=\textwidth]{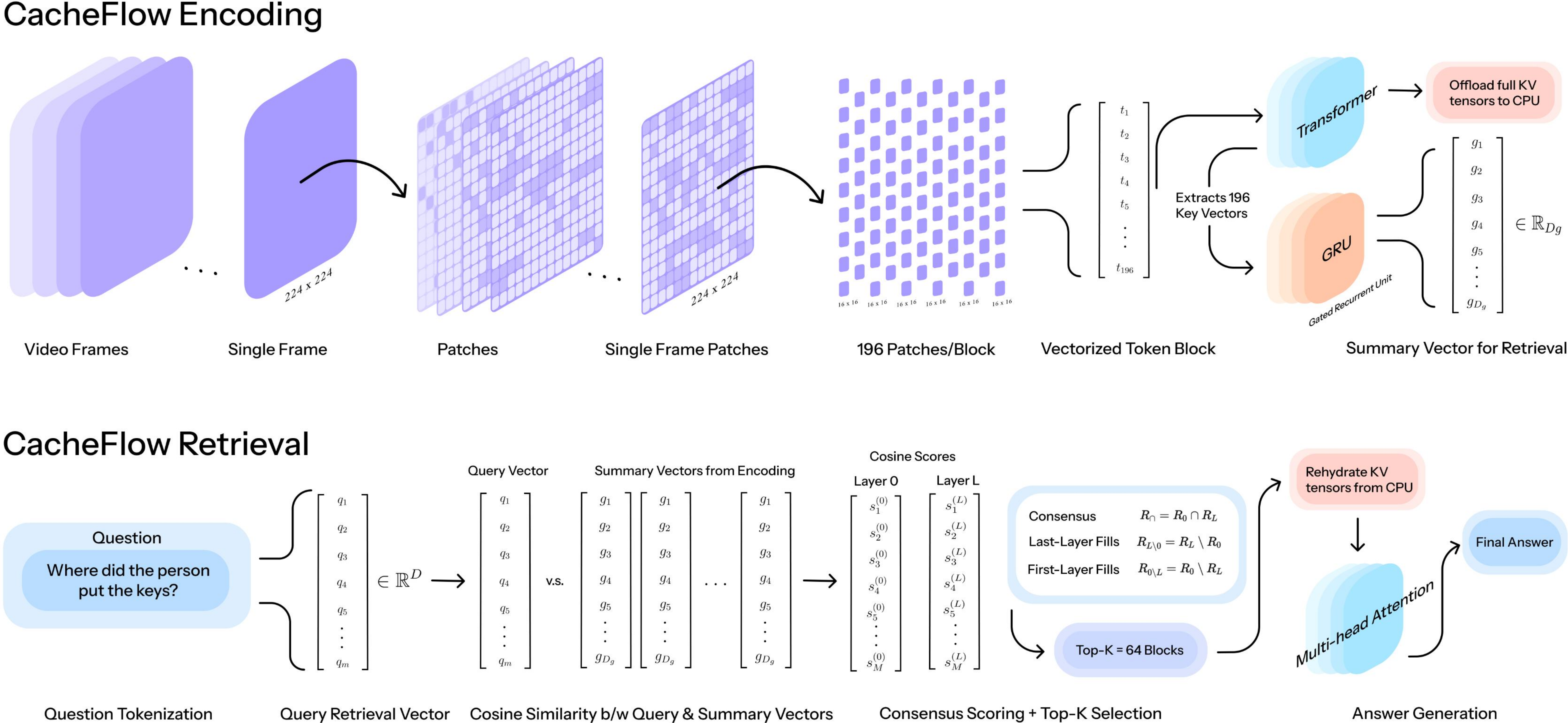} 
  \caption{\textbf{CacheFlow system overview.} Dynamic Token Dropping prunes redundant tokens via inter-frame similarity, and surviving tokens are packed into fixed-size blocks. Each block is summarized by a GRU-based compressive memory, while full key–value pairs are offloaded. During inference, consensus-first retrieval rehydrates only the Top-$K$ relevant blocks for efficient long-range reasoning.}
  \label{fig:cacheflow_system}
  \vspace{-0.75em}
\end{figure*}

\begin{abstract}
Long-form video question answering (VQA) overwhelms current vision-language models (VLMs) because attention and key-value (KV) caches grow with runtime, forcing either expensive inference or near-sighted sliding windows. We introduce CacheFlow, a training-free pipeline that pairs Dynamic Token Dropping (DTD) with a compressive long-term memory. DTD prunes per-patch tokens online via cosine similarity to the previous frame, and surviving tokens are packed into fixed-size blocks. This online, per-frame processing makes our approach fundamentally suited for live streaming VQA. As blocks are processed, each one’s keys are summarized by a tiny recurrent encoder to form a retrieval index, while the block’s full KV pairs are offloaded and later rehydrated for generation, preserving answer fidelity. At inference, a consensus-based retrieval mechanism retrieves only the Top-K most relevant blocks and attends over both the retrieved and local context for precise, long-range reasoning. CacheFlow is drop-in, architecture-agnostic, and requires no fine-tuning. Experiments on both offline and streaming VQA benchmarks demonstrate that CacheFlow outperforms current strong baselines, while processing up to ~87\% less tokens. Our dual approach enables VLMs to be both efficient and context-aware, paving the way for practical long-form video understanding.
\end{abstract}

\section{Introduction}

Long-form video understanding presents a fundamental challenge for modern vision-language models (VLMs). While recent multimodal architectures such as LLaVA and Qwen-VL \cite{yang2024qwen2} have achieved remarkable progress on short clips, their attention-based mechanisms scale quadratically, and for videos spanning minutes or hours the cumulative key–value (KV) cache quickly exceeds GPU memory, forcing models to either truncate context or perform expensive recomputation. This leaves a persistent gap between VLM efficiency and the temporal complexity of real-world video. 

Existing approaches to long-term modeling provide only partial solutions. Sliding-window and recurrent attention methods preserve computational tractability but inherently discard distant context once older tokens fall outside the window, limiting temporal reasoning to short horizons. Retrieval-augmented and memory-based systems increase the effective context length, yet they often accumulate redundant frame-level features and depend on uniform pooling or static compression that blur fine-grained temporal structure. These methods either underutilize memory by repeatedly storing similar information or overcompress it by collapsing temporally distinct events into coarse averages. Consequently, current designs fail to achieve both efficiency and fidelity—maintaining compact memory representations without sacrificing precise, temporally grounded recall.

We introduce \textbf{CacheFlow}, a training-free framework that bridges this gap through a two-pronged design: \textbf{Dynamic Token Dropping (DTD)} and \textbf{GRU-based Compressive Memory (GCM)}. DTD reduces redundancy at the source by adaptively pruning per-patch tokens whose features remain similar across frames, yielding a sparse yet information-rich token stream. The GCM module complements this by compressing each block’s keys through a lightweight recurrent encoder, producing temporally-aware summaries that enable fast retrieval while preserving semantic integrity. During inference, a \textit{consensus-first retrieval} strategy selectively rehydrates only the most relevant cached blocks, allowing the model to attend jointly to both recent and retrieved context. This design achieves long-range temporal reasoning without overwhelming the KV cache or requiring model retraining. 

CacheFlow operates online, requires no fine-tuning, and integrates seamlessly with existing VLMs. Across multiple long-form and streaming VQA benchmarks, it consistently improves both efficiency and accuracy, \textbf{reducing KV cache size by up to 87\%} while maintaining \textbf{state-of-the-art} performance. With dynamic token pruning and compressive memory, CacheFlow establishes a scalable foundation for practical, streaming-capable vision-language systems.

\section{Related Work}

\subsection{Efficient Transformers for Vision}

Transformers have become the dominant backbone for vision tasks, yet their quadratic attention cost poses major challenges for scalability. A broad line of work seeks to improve efficiency by constraining the attention pattern or reducing token count. Hierarchical and window-based variants such as Swin Transformer~\cite{liu2021swin} and Longformer~\cite{beltagy2020longformer} limit pairwise interactions to local neighborhoods, achieving subquadratic complexity at the expense of long-range dependencies. Linearized approximations including Performer~\cite{choromanski2021performer} and Linformer~\cite{wang2020linformer} instead reparameterize the attention kernel to achieve linear time, but often degrade accuracy on dense spatial reasoning tasks. Complementary strategies such as token merging or pruning—exemplified by ToMe~\cite{bolya2023tome} and H-ViT~\cite{Ghahremani24HViT}—fuse redundant patches to accelerate inference while maintaining visual coverage. Although these advances substantially reduce compute for individual images or short clips, they do not address the growing key–value (KV) cache that accumulates when processing long video streams, where redundancy arises not only spatially but also temporally.

\subsection{Long-Context Video Understanding}

Modeling extended temporal context in video understanding remains an open frontier.  
Conventional sliding-window and recurrent attention architectures~\cite{bertasius2021timesformer,liu2022videoswin,arnab2021vivit,fan2021mvit} maintain efficiency by retaining only recent tokens, but this truncation inherently limits long-range reasoning. Retrieval-augmented and persistent-memory systems~\cite{sun2023retentive,lewis2020retrieval,wu2022memvit,Xiao2023EfficientStreamingLanguageModels} attempt to bridge this gap by storing and reusing intermediate representations, yet often at the expense of redundancy or with retrieval mechanisms too coarse to capture nuanced temporal cues. ReKV~\cite{Di2025Streaming} introduced retrieval over stored KV-caches for streaming video reasoning, but its block-level design still produces overlapping storage and reduced temporal precision. Flash-VStream~\cite{ZhangEtAl2024FlashVStream} advanced efficiency further through a hierarchical flash memory with synopsis and detail layers, enabling real-time inference but emphasizing throughput over representational compression.  
\textbf{CacheFlow} unifies temporal sparsity and compressive memory within the transformer architecture, offering a new pathway toward \textbf{efficient long-term video reasoning}. Through adaptive token pruning and a recurrent compressive cache, it maintains fine-grained temporal fidelity while adhering to tight memory and latency budgets.

\subsection{Temporal Redundancy in Videos}

Natural videos exhibit substantial temporal redundancy, where consecutive frames differ only in minor spatial details.  
Early work addressed this redundancy through heuristic frame sampling~\cite{Wu2019LFB,Jiang2020DivideAndConquer} or keyframe selection~\cite{JadonJasim2019}, reducing computation at the cost of temporal continuity. Subsequent methods introduced motion-based or feature-level filtering~\cite{zhang2022motionbert,rao2021dynamicvit} to discard visually similar regions across frames. More recently, TimeChat~\cite{timechatonline} proposed differential token dropping based on inter-frame cosine similarity, showing that many patch-level features remain static over time. While effective for streaming efficiency, it does not maintain long-term contextual coherence once dropped tokens are removed from memory. CacheFlow extends this idea by pairing online, \textbf{per-frame token pruning} with a \textbf{compressive recurrent memory} that stores compact summaries of past context, achieving temporal sparsity without losing access to long-range dependencies.

\section{Method}

We present \textbf{CacheFlow}, a training-free pipeline for long-video understanding built around two ideas: (i) \textbf{Dynamic Token Dropping (DTD)} to eliminate frame-to-frame redundancy at the source and (ii) a \textbf{GRU-based Compressive Memory (GCM)} to compress, store, and selectively reload long-range context as needed for answering.

Figure \ref{fig:cacheflow_system} provides an end-to-end visual overview of the CacheFlow architecture, illustrating how Dynamic Token Dropping, GRU-based Compressive Memory, and consensus-based retrieval interact during inference.


\subsection{System Overview}

Let a video be a sequence of frames, which the vision tower maps into a set of patch features for each frame $X_t = \{x_{t,p}\}_{p=1}^P$.  
Our DTD module computes a binary ``keep'' mask $m_{t,p} \in \{0, 1\}$ for each patch by comparing its feature vector $x_{t,p}$ to the corresponding patch in the prior frame, producing a pruned set of tokens $X'_t = \{x_{t,p} \mid m_{t,p}=1\}$.

These surviving tokens are accumulated across frames and dynamically packed into fixed-size blocks of $B$ tokens (e.g., $B{=}196$), denoted $\mathcal{B}_j$.  
These blocks flow through the language model, which generates per-layer keys and values $(K^\ell, V^\ell)$.  
A cache manager maintains a local window of the most recent blocks on GPU.  
When a block $\mathcal{B}_j$ is offloaded, its layer-wise Key tensor $K^\ell_j \in \mathbb{R}^{H \times B \times D_k}$ (where $H$ is the number of heads and $D_k$ is the key dimension) is compressed by a GRU into a single hidden state
\begin{equation}
g^\ell_j = \text{GRU}_\theta(\text{reshape}(K^\ell_j)) \in \mathbb{R}^{D_g},
\end{equation}
which acts as its retrieval key.  
The block's raw $(K^\ell, V^\ell)$ pairs are offloaded to CPU memory.

At question time, a two-pass retrieval process occurs.  
First, the question is used to retrieve the Top-$K$ global blocks using a \textit{consensus-first} algorithm.  
Second, only these $K$ blocks are ``rehydrated'' to the GPU, their positional embeddings are correctly offset, and the model performs the final answer-generation pass by attending over this combined retrieved and local context.

\subsection{Dynamic Token Dropping (DTD)}

Given the feature vectors $x_{t,p}$ for a patch $p$ at frame $t$, we compute a redundancy score using cosine similarity against the previous frame:
\begin{equation}
s_{t,p} = \cos(x_{t,p}, x_{t-1,p}) = 
\frac{\langle x_{t,p}, x_{t-1,p}\rangle}{\|x_{t,p}\|\|x_{t-1,p}\|}.
\end{equation}
We drop tokens that are too similar to the previous frame by applying a mask $m_{t,p}$ governed by a threshold $\tau_{\text{feat}}$
\begin{equation}
m_{t,p} = \mathds{1}[s_{t,p} < \tau_{\text{feat}}].
\end{equation}

To ensure stability, the first frame of any video ($t{=}1$) is always kept in its entirety to establish a clean reference.  
Furthermore, if all tokens in a frame are masked out (i.e., $\sum_p m_{t,p} = 0$), we force-keep the token with the lowest similarity to prevent empty frames.  
This DTD process significantly reduces the total number of tokens processed, lowering both the computational cost of attention and the memory footprint of the KV cache. 

\subsection{Dynamic Token-to-Block Packing}

The pruned, variable-length token sets \( X'_t \) from each frame are concatenated into a single temporal stream.
\begin{equation}
\mathcal{S} = \bigcup_{t=1}^{T} X'_t,
\qquad
\mathcal{B}_j = \mathcal{S}[(j{-}1)B+1:jB],
\end{equation}
where $T$ is the number of frames and $\mathcal{B}_j$ denotes the $j$-th full block.  
If the stream ends with an incomplete residual block, it is either padded to size $B$ (if it is the only block) or discarded to ensure the cache manager processes only uniform blocks. Uniform block structure stabilizes cache eviction and retrieval, allowing CacheFlow to operate in a streaming fashion without irregular memory fragmentation.

\subsection{GRU-based Compressive Memory}

When a fixed-size block $\mathcal{B}_j$ is processed and must be evicted from the GPU's local window, its Key and Value tensors $(K^\ell_j, V^\ell_j)$ at each layer $\ell$ are offloaded.

\subsubsection{GRU Compressor}

Instead of simple averaging, we generate a representative vector by treating the block's Key tensor as a sequence.  
The tensor $K^\ell_j$ is reshaped into a sequence of $B$ vectors $k'_{j,i} \in \mathbb{R}^{H\!\cdot\!D_k}$ and passed through a lightweight, single-layer GRU:
\begin{align}
r_i &= \sigma(W_r k'_{j,i} + U_r h_{i-1} + b_r), \\
z_i &= \sigma(W_z k'_{j,i} + U_z h_{i-1} + b_z), \\
\tilde{h}_i &= \tanh(W_h k'_{j,i} + U_h (r_i \odot h_{i-1}) + b_h), \\
h_i &= (1-z_i) \odot h_{i-1} + z_i \odot \tilde{h}_i.
\end{align}

The GRU operates in a \textit{frozen, training-free} manner: its parameters remain fixed during inference and serve purely as a deterministic, recurrent summarizer. Even without learning, its gating dynamics provide a strong temporal inductive bias, enabling it to accumulate salient variations across the sequence while discarding redundant information. Empirically, this fixed recurrent summarization yields more stable and order-aware representations than mean pooling or linear projection, while requiring no additional training. Detailed quantitative comparisons are provided in Section~\ref{sec:results}.

The final hidden state $g^\ell_j = h_B$ serves as the compressed retrieval key for the entire block and is stored on GPU with a pointer to the full $(K^\ell_j, V^\ell_j)$ on CPU.  
Across all blocks, these vectors form a compact index $G^\ell = [g^\ell_1, \ldots, g^\ell_M]$ used for similarity-based retrieval.

\subsubsection{Consensus-First Retrieval}

During inference, the question is first passed through the model to generate query vectors $q^\ell$ at each layer.  
We then perform a retrieval pass by computing cosine similarity between each query and stored block key:
\begin{equation}
s^\ell_j = \frac{q^\ell \cdot g^\ell_j}{\|q^\ell\| \|g^\ell_j\|}.
\end{equation}
Top-$K$ indices from the first ($\mathcal{R}_0$) and last ($\mathcal{R}_L$) layers are combined to form the final set $\mathcal{R}_{\text{final}}$ as follows:
\begin{enumerate}
\item Summing blocks in $\mathcal{R}_0 \cap \mathcal{R}_L$ (consensus).
\item Filling remaining slots with the highest-scoring blocks from $\mathcal{R}_L \setminus \mathcal{R}_0$ (LN-fill).
\item If needed, adding leftover blocks from $\mathcal{R}_0 \setminus \mathcal{R}_L$ (L0-fill)
until $|\mathcal{R}_{\text{final}}|{=}K$.
\end{enumerate}

\vspace{2pt}

\noindent Formally,
\begin{equation}
\mathcal{R}_{\text{final}} =
(\mathcal{R}_0 \cap \mathcal{R}_L)
\cup
(\mathcal{R}_L \setminus \mathcal{R}_0)
\cup
(\mathcal{R}_0 \setminus \mathcal{R}_L),
\end{equation}
truncated to the top-$K$ entries.  
This consensus-first approach prioritizes blocks judged salient by both shallow and deep layers, improving retrieval robustness.

\subsubsection{Position-Aware Attention (RoPE with Offset)}

When the retrieved blocks are rehydrated to GPU memory, they are prepended to the local window.  
To preserve temporal ordering, the text prompt's tokens must have their rotary positions offset.  
If the retrieved context length is $n_{\text{mem}}$, the query tokens with relative indices $i{=}1,\ldots,n_q$ are assigned absolute positions $[n_{\text{mem}}{+}1, \ldots, n_{\text{mem}}{+}n_q]$, ensuring continuous rotational phases across cached and new positions.

\subsection{Answering with Retrieved Memory}

Let $Q^\ell$ be the query projections from the question, and let $K^\ell_{\text{cache}}$ and $V^\ell_{\text{cache}}$ denote the concatenated keys and values from both the retrieved (global) and local (window) blocks after position correction.  
The attention computation is then
\begin{equation}
\text{Attn}(Q^\ell, K^\ell_{\text{cache}}, V^\ell_{\text{cache}}) 
= 
\text{softmax}\!\left(\frac{Q^\ell (K^\ell_{\text{cache}})^\top}{\sqrt{d_k}}\right)
V^\ell_{\text{cache}}.
\end{equation}

Because attention is now bounded by the local window size and the $K$ retrieved blocks, its complexity scales as
\begin{equation}
\mathcal{O}\big((n_{\text{local}} + K B)^2\big) 
\quad \text{instead of} \quad 
\mathcal{O}(T^2),
\end{equation}
where $T$ is the total video length.  
CacheFlow can thus process long videos efficiently while accurately retrieving and attending to fine-grained, long-range dependencies. Together, these components form a fully streaming, drop-in memory architecture that preserves long-range reasoning while remaining lightweight enough for real-time deployment.

\section{Experiments}

We evaluate CacheFlow on long-form and streaming video question answering (VQA) tasks, focusing on scenarios where temporal context spans minutes and conventional key–value caches become prohibitively large.

\subsection{Datasets and Metrics}

\noindent We conduct experiments across a diverse set of datasets which span egocentric, cinematic, and open-domain settings, covering both multiple-choice and free-form QA tasks.  
Together, they assess temporal reasoning, retrieval precision, and scalability under offline and streaming conditions.

\noindent\textbf{QAEgo4D}~\cite{barmann2022qaego4d} builds on the large-scale Ego4D~\cite{grauman2022ego4dworld3000hours} dataset of first-person videos.  
It contains natural question--answer pairs grounded in egocentric activities and serves as a testbed for evaluating temporal reasoning and object--interaction understanding over continuous video segments.

\noindent\textbf{EgoSchema}~\cite{mangalam2023egoschema} extends the egocentric paradigm to multi-minute clips with 5-way multiple-choice questions requiring cross-scene and causal reasoning.  
It explicitly measures a model's ability to retain information and reason coherently across long temporal horizons.

\noindent\textbf{ActivityNet-QA}~\cite{yu2019activitynetqa} consists of open-domain videos covering diverse human activities with free-form natural-language answers.  
Its wide coverage and sentence-level queries test compositional reasoning and general-purpose video understanding.

\noindent\textbf{MLVU}~\cite{zhou2024mlvu} benchmark evaluates multi-task, multi-duration long-video comprehension across nine tasks.  
It includes 1{,}122 videos averaging 14 minutes each, drawn from movies, documentaries, egocentric recordings, and gameplay footage, supporting hierarchical reasoning across multiple temporal scales.

\noindent\textbf{RVS-Ego} and \textbf{RVS-Movie}~\cite{ivgsz2024rvsego, ivgsz2024rvsmovie} are retrieval-based long-video benchmarks from egocentric and cinematic domains, respectively.  
They require models to identify temporally distant visual evidence relevant to textual prompts, making them ideal for stress-testing CacheFlow's retrieval and long-range reasoning capabilities.

\paragraph{Metrics.}  
We evaluate CacheFlow across four primary dimensions: accuracy, speed, memory usage, and cache efficiency.  
For multiple-choice benchmarks, we report \textbf{answer accuracy}, and for open-ended datasets, we follow prior work in using the \texttt{gpt-3.5-turbo} model as an automatic judge to assess answer correctness.  
The judging prompt is consistent with those used in prior work and is provided as supplementary material for completeness.  
To measure efficiency, we track \textbf{latency} (in seconds) for answer generation time as well as the token drop percentage (applicable for our CacheFlow method).  
We also measure \textbf{peak GPU memory consumption} in gigabytes to quantify resource usage during inference.  
These metrics help in capturing the effectiveness of CacheFlow’s compressive streaming memory in reducing long-term attention state without compromising accuracy.

\subsection{Baselines}

\noindent We evaluate CacheFlow against a representative set of state-of-the-art video–language models and long-context architectures spanning both proprietary and open-source systems.  
Commercial multimodal models such as \textbf{GPT-4V} and \textbf{Gemini 2.5 Pro} serve as upper-bound references for fully trained, large-scale systems.  
For open-source baselines, we include \textbf{LongVA-7B} and \textbf{Flash V-Stream-7B}, which implement learned compression and hierarchical temporal aggregation for long-video reasoning.  
To ensure a controlled comparison, both \textbf{ReKV} and \textbf{CacheFlow} are applied on the same \textbf{LLaVA-OneVision} backbones at two scales (0.5B and 7B parameters), enabling direct evaluation of their respective memory mechanisms under identical model capacities. We also include the vanilla performance of both versions of LlaVA-OneVision models as a baseline.
All methods are tested using the same inference conditions, and prompt templates, providing a unified benchmark methodology for efficiency, memory usage, and reasoning performance.

\subsection{Implementation Details}

We implement CacheFlow on both \textbf{LLaVA-OV-0.5B} and \textbf{LLaVA-OV-7B} backbones~\cite{li2024llavaonevision}, matching the model scales used in all offline and streaming evaluations. Each video frame is decomposed into \(14\times 14\) patches. After applying Dynamic Token Dropping (DTD), the surviving tokens are reassembled into fixed-size blocks of \(B{=}196\) tokens. The cache maintains the most recent \(n_{\text{local}}\) blocks on GPU, while older blocks are offloaded and compressed. Unless otherwise specified, we retrieve \(K{=}64\) blocks per query using the \emph{consensus-first} strategy that aggregates retrieval signals from both the first and last transformer layers. This is to maintain consistency with prior work. All experiments, offline and streaming, are run on a single NVIDIA A100 GPU to ensure consistent evaluation conditions.

\section{Results and Analysis}
\label{sec:results}

We evaluate CacheFlow on both \emph{offline} and \emph{streaming} video question-answering benchmarks. Our primary comparisons are against (i) state-of-the-art open-source long-video systems (LongVA-7B, Flash V-Stream-7B) and (ii) the training-free ReKV baseline instantiated on LLaVA-OneVision backbones at both 0.5B and 7B scales. For completeness, we also report (iii) strong commercial multimodal models (GPT-4V, Gemini 2.5 Pro), whose proprietary training and scale make them unsuitable as direct baselines. 
Unless otherwise noted, all ReKV and CacheFlow results share the same backbone and inference configuration. CacheFlow is applied as a \textbf{training-free, drop-in replacement} for ReKV’s memory module.

\subsection{Offline long-video QA}
\label{sec:offline_results}

\begin{table*}[t]
\centering
\setlength{\tabcolsep}{13pt}
\caption{\textbf{Offline video QA comparison.} For QAEgo4D, MLVU, and EgoSchema, we report accuracy (\%) and token-drop (\%) when evaluating CacheFlow. For ActivityNet-QA, we report an additional score which denotes the open-ended answer quality rating produced by \texttt{gpt-3.5-turbo}.}
\resizebox{\textwidth}{!}{
\begin{tabular}{lccccccccc}
\toprule
\multirow{2}{*}{Method}
& \multicolumn{2}{c}{QAEgo4D}
& \multicolumn{2}{c}{MLVU}
& \multicolumn{2}{c}{EgoSchema}
& \multicolumn{3}{c}{ActivityNet-QA} \\
\cmidrule(lr){2-3}
\cmidrule(lr){4-5}
\cmidrule(lr){6-7}
\cmidrule(lr){8-10}
& Acc. & Drop 
& Acc. & Drop 
& Acc. & Drop 
& Acc. & Score & Drop \\
\midrule
GPT-4V \cite{openai2023gpt4v_system_card}               & --   & --    & 49.2 & --    & --   & --    & 57.0   & --    & --    \\
Gemini 2.5 Pro \cite{deepmind2025gemini25}         & --   & --    & --   & --    & --   & --    & 66.7 & --    & --    \\
LongVA-7B \cite{zhang2024long}            & --   & --    & 56.3 & --    & --   & --    & 50.0   & 3.30   & --    \\
Flash V-Stream-7B \cite{ZhangEtAl2024FlashVStream}      & 38.2 & --    & 50.2 & --    & 38.1 & --    & 51.9 & 3.40   & --    \\
\midrule
LLaVA-OV-0.5B         & 42.6 & --    & 53.2 & --    & 26.6 & --    & 50.5 & 3.02  & --    \\
\rowcolor{gray!10}
\quad ReKV                      & 48.2 & --    & 54.2 & --    & 25.6 & --    & 49.9 & 3.01  & --    \\
\rowcolor{gray!10}
\quad \textbf{CacheFlow ($\tau{=}0.5$)} 
                                & \textbf{50.8} & \textbf{81.8}
                                & \textbf{54.6} & \textbf{79.7}
                                & \textbf{27.9} & \textbf{82.6}
                                & \textbf{50.6} & \textbf{3.47}
                                & \textbf{80.8} \\
\midrule
LLaVA-OV-7B            & 52.8 & --    & 64.7 & --    & 59.8 & --    & 56.6 & 3.29  & --    \\
\rowcolor{gray!10}
\quad ReKV                      & 54.8 & --    & 66.5 & --    & 60.7 & --    & 60.2 & 3.52  & --    \\
\rowcolor{gray!10}
\quad \textbf{CacheFlow ($\tau{=}0.5$)} 
                                & \textbf{55.4} & \textbf{70.9}
                                & \textbf{66.9} & \textbf{67.1}
                                & \textbf{60.2} & \textbf{72.0}
                                & \textbf{59.6} & \textbf{3.64}
                                & \textbf{70.9} \\
\bottomrule
\end{tabular}}
\label{tab:cross_benchmark_results}
\vspace{-0.5em}
\end{table*}

Table~\ref{tab:cross_benchmark_results} summarizes offline results on QAEgo4D, MLVU, EgoSchema, and ActivityNet-QA.
We report accuracy for all methods, and additionally report token-drop ratios (percentage of visual tokens pruned by Dynamic Token Dropping) for CacheFlow.

\paragraph{CacheFlow performance at 0.5B scale.}
On the lighter LLaVA-OV-0.5B backbone, CacheFlow consistently improves over the ReKV baseline while discarding the vast majority of visual tokens.
On QAEgo4D, accuracy increases from 48.2\% (ReKV) to 50.8\% with CacheFlow, a +2.6 point absolute gain while dropping 81.8\% of tokens.
On MLVU, CacheFlow slightly improves accuracy from 54.2\% to 54.6\% with 79.7\% token-drop.
The gains are largest on EgoSchema: accuracy rises from 25.6\% to 27.9\% (+2.3 points) despite removing 82.6\% of visual tokens.
On ActivityNet-QA, CacheFlow improves both overall accuracy (49.9\% $\rightarrow$ 50.6\%) and open-ended answer score (3.01 $\rightarrow$ 3.47) while dropping 80.8\% of tokens.
Taken together, these results indicate that under the same 0.5B backbone, CacheFlow \textbf{not only preserves but often \emph{improves}} QA performance relative to ReKV, even when it discards roughly 80\% of all video tokens.

\paragraph{CacheFlow performance at 7B scale.}
On the stronger 7B backbone, we observe similar trends.
CacheFlow boosts QAEgo4D accuracy from 54.8\% to 55.4\% and MLVU from 66.5\% to 66.9\%, while dropping 70.9\% and 67.1\% of tokens, respectively.
On EgoSchema, CacheFlow \textbf{remains competitive} (60.2\% vs.\ 60.7\% for ReKV) despite removing 72.0\% of tokens.
For ActivityNet-QA, CacheFlow achieves a slightly lower overall accuracy (59.6\% vs.\ 60.2\% for ReKV) but \emph{improves} the LLM-based response rating (3.52 $\rightarrow$ 3.64), while still discarding around 71\% of tokens.
These results show that CacheFlow can be plugged into a stronger backbone without re-tuning for \textbf{extremely efficient} inference: it matches or slightly exceeds ReKV in accuracy on three out of four benchmarks tested and improves open-ended response quality on ActivityNet-QA, while aggressively reducing the effective visual context size.

\paragraph{Comparison to commercial and open-source long-video models.}
The top block of Table~\ref{tab:cross_benchmark_results} situates our training-free method against commercial MLLMs and specialized open-source long-video models.
On ActivityNet-QA, CacheFlow with LLaVA-OV-7B reaches 59.6\% accuracy and a 3.64 LLM response score, substantially narrowing the gap to GPT-4V (57.0\%) and Gemini 2.5 Pro (66.7\%) while using \textit{only} an open-source backbone.
Compared to LongVA-7B and Flash V-Stream-7B, CacheFlow-equipped LLaVA-OV-7B provides \textbf{stronger or comparable accuracy} across the offline benchmarks, despite not relying on custom long-context pretraining.
These comparisons underscore that training-free compression and retrieval can close much of the performance gap to heavily engineered or proprietary systems, provided that temporal redundancy is handled carefully.

\subsection{Streaming video QA}
\label{sec:streaming_results}

\begin{table*}[h!]
\centering
\setlength{\tabcolsep}{12pt}
\caption{\textbf{Streaming video QA on RVS.} For RVS-Ego and RVS-Movie, we report accuracy (\%), score, and token-drop (\%) when evaluating CacheFlow. \textit{Score} denotes the open-ended answer quality rating produced by \texttt{gpt-3.5-turbo}. \textit{Latency} is the wall-clock inference time per example (s), and \textit{GPU Mem} is peak GPU memory usage during evaluation (GB).}
\resizebox{\textwidth}{!}{
\begin{tabular}{lcccccccc}
\toprule
\multirow{2}{*}{Method}
& \multicolumn{3}{c}{RVS-Ego}
& \multicolumn{3}{c}{RVS-Movie}
& \multicolumn{2}{c}{Running Metrics} \\
\cmidrule(lr){2-4}
\cmidrule(lr){5-7}
\cmidrule(lr){8-9}
& Acc. & Score & Drop 
& Acc. & Score & Drop 
& Latency & GPU Mem \\
\midrule
Flash V-Stream-7B  & 56.3 & 3.90 & --   & 53.3 & 3.30 & --    & 2.70s & 26 GB   \\
\midrule
LLaVA-OV-0.5B\\
\rowcolor{gray!10}
\quad ReKV         & 51.9 & 3.80 & --   & 42.9 & 3.34 & --    & 1.60s & 19 GB   \\
\rowcolor{gray!10}
\quad \textbf{CacheFlow ($\tau{=}0.5$)} 
                   & \textbf{54.3} & \textbf{3.87} & \textbf{86.8}
                   & \textbf{42.6} & \textbf{3.34} & \textbf{70.7}
                   & \textbf{1.38s} & \textbf{2.9 GB} \\
\midrule
LLaVA-OV-7B \\
\rowcolor{gray!10}
\quad ReKV         & 59.7 & 3.90 & --   & 49.6 & 3.40 & --    & 3.30s & 38 GB   \\
\rowcolor{gray!10}
\quad \textbf{CacheFlow ($\tau{=}0.5$)} 
                   & \textbf{61.6} & \textbf{3.94} & \textbf{78.5}
                   & \textbf{50.5} & \textbf{3.44} & \textbf{58.0}
                   & \textbf{2.03s} & \textbf{26.4 GB} \\
\bottomrule
\end{tabular}}
\label{tab:rvs_streaming_results}
\vspace{-0.5em}
\end{table*}

Table~\ref{tab:rvs_streaming_results} evaluates CacheFlow in the \emph{streaming} regime on RVS-Ego and RVS-Movie, where videos are consumed sequentially and the model must answer questions online while respecting latency and memory constraints.

\paragraph{Streaming with a 0.5B backbone.}
On LLaVA-OV-0.5B, CacheFlow improves RVS-Ego accuracy from 51.9\% (ReKV) to 54.3\% (+2.4 points) and increases the LLM response rating from 3.80 to 3.87, while discarding 86.8\% of tokens.
On RVS-Movie, CacheFlow matches ReKV in answer quality with a small change in accuracy (42.9\% to 42.6\%), despite dropping 70.7\% of tokens.
Crucially, these accuracy levels are achieved with \emph{better} efficiency: latency is reduced from 1.60\,s to 1.38\,s per example ($\sim$13.8\% reduction), and peak GPU memory drops from 19\,GB to 2.9\,GB, a $\sim$6.6$\times$ reduction.
Thus, at small scale, CacheFlow provides \textbf{strictly better accuracy--efficiency} trade-offs for ego-centric streaming, and preserves performance for movie-style content while dramatically reducing the runtime footprint.

\paragraph{Streaming with a 7B backbone.}
At 7B scale, the benefits of CacheFlow are \textit{even more} pronounced.
On RVS-Ego, CacheFlow raises accuracy from 59.7\% to 61.6\% (+1.9 points) and slightly improves the LLM response rating (3.90 $\rightarrow$ 3.94), while dropping 78.5\% of tokens. On RVS-Movie, accuracy increases from 49.6\% to 50.5\% (+0.9 points), with a score gain from 3.40 to 3.44 and 58.0\% token-drop.
Despite these gains, CacheFlow reduces latency from 3.30\,s to 2.03\,s per example ($\sim$38\% reduction) and lowers peak GPU usage from 38\,GB to 26.4\,GB ($\sim$1.4$\times$ memory savings).
Compared to the Flash V-Stream-7B streaming baseline, CacheFlow-equipped LLaVA-OV-7B attains higher accuracy on both RVS-Ego (61.6\% vs.\ 56.3\%) and competitive accuracy on RVS-Movie (50.5\% vs.\ 53.3\%, but with higher answer quality scores), while operating with competitive or better latency and memory.
These results demonstrate that CacheFlow \textbf{scales favorably} to multi-minute, open-domain streams and can be used to make larger backbones practical in streaming settings.

\subsection{Ablation studies}
\label{sec:ablations}

We next isolate the contributions of the Dynamic Token Dropping (DTD) module and the GRU-based compressive controller.
All ablations are conducted using the same backbones and evaluation protocols as in the main experiments.

\subsubsection{Effect of DTD threshold}
\label{sec:ablation_dtd}

Table~\ref{tab:ablate_tau} studies how the DTD threshold \(\tau_{\text{feat}}\) controls the trade-off between compression and accuracy on MLVU and RVS-Ego.
A lower threshold (0.25) is highly aggressive: it removes 98.7\% of tokens on MLVU and 99.3\% on RVS-Ego.
This extreme pruning degrades accuracy (52.1\% on MLVU, 52.8\% on RVS-Ego) relative to more moderate thresholds.
Increasing \(\tau_{\text{feat}}\) to \textbf{0.50 relaxes the pruning and yields a better balance}: accuracy improves to 54.6\% on MLVU and 54.3\% on RVS-Ego, while still discarding 79.7\% and 86.8\% of tokens, respectively. Further increasing \(\tau_{\text{feat}}\) to 0.75 reduces compression (only 35.9\% and 45.8\% token-drop) and slightly boosts RVS-Ego accuracy (55.4\%) and score (3.90), but yields a lower MLVU accuracy (53.4\%) compared to \(\tau_{\text{feat}}{=}0.5\).

\begin{table}[h!]
\centering
\setlength{\tabcolsep}{8pt}
\caption{\textbf{Effect of DTD threshold \(\tau_{\text{feat}}\) on MLVU and RVS-Ego.}
Lower values yield stronger compression; higher values retain more visual tokens.}
\begin{tabular}{lccccc}
\toprule
\multirow{2}{*}{\(\tau_{\text{feat}}\)}
& \multicolumn{2}{c}{MLVU}
& \multicolumn{3}{c}{RVS-Ego} \\
\cmidrule(lr){2-3}
\cmidrule(lr){4-6}
& Acc. & Drop 
& Acc. & Score & Drop \\
\midrule
0.25 & 52.1 & 98.7 & 52.8 & 3.85 & 99.3 \\
\rowcolor{gray!10}
0.50 & 54.6 & 79.7 & 54.3 & 3.87 & 86.8 \\
0.75 & 53.4 & 35.9 & 55.4 & 3.90 & 45.8 \\
\bottomrule
\end{tabular}
\label{tab:ablate_tau}
\vspace{-0.4em}
\end{table}

These trends indicate that most \textbf{inter-frame information is indeed redundant}, and that substantial token-dropping is possible without catastrophic performance loss.
However, overly aggressive thresholds (e.g., 0.25) can discard genuinely informative frames, while overly conservative thresholds (e.g., 0.75) sacrifice efficiency benefits.
In practice, \(\tau_{\text{feat}}{=}0.5\) emerges as a robust operating point that offers strong accuracy across both MLVU and RVS-Ego, with 80--87\% compression.

\subsubsection{Mean pooling vs.\ GRU compression}
\label{sec:ablation_gru}

\noindent Table~\ref{tab:ablate_gru} compares two ways of compressing past visual tokens when DTD is disabled: a simple mean-pooling baseline and the GRU-based controller used in CacheFlow.
On QAEgo4D, replacing mean pooling with the GRU increases accuracy from 47.6\% to 48.2\%.
On RVS-Ego, the GRU variant improves accuracy from 51.9\% to 52.9\% and slightly raises the LLM response rating from 3.80 to 3.85.
Although the absolute gains appear modest, they are achieved \emph{without} any additional training or changes to the backbone, and they \textbf{consistently favor the GRU} across both offline and streaming settings.

\begin{table}[h!]
\centering
\setlength{\tabcolsep}{8pt}
\caption{\textbf{Ablation of GRU controller on CacheFlow.}
Both variants disable DTD, and the GRU controller yields modest improvements on QAEgo4D (offline) and RVS-Ego (streaming).}
\begin{tabular}{lccc}
\toprule
\multirow{2}{*}{%
  \begin{tabular}{@{}c@{}}
    CacheFlow\\
    {\footnotesize\textit{w/o DTD}}
  \end{tabular}
}
& \multicolumn{1}{c}{QAEgo4D}
& \multicolumn{2}{c}{RVS-Ego} \\
\cmidrule(lr){2-2}
\cmidrule(lr){3-4}
& Acc. & Acc. & Score \\
\midrule
Mean Pooling & 47.6 & 51.9 & 3.80 \\
\rowcolor{gray!10}
GRU          & 48.2 & 52.9 & 3.85 \\
\bottomrule
\end{tabular}
\label{tab:ablate_gru}
\vspace{-0.4em}
\end{table}

These results suggest that even a frozen, lightweight recurrent controller is preferable to static averaging for long-range summarization.
By maintaining an order-sensitive hidden state, the GRU can better preserve temporal dependencies and subtle event structure, which later retrieval and reasoning stages can exploit.
Combined with DTD, this recurrent summarization is a key ingredient that enables CacheFlow to recover or exceed ReKV-level accuracy while operating at much higher compression rates on both offline and streaming long-video QA benchmarks.

\subsubsection{Effect of Consensus-Based Retrieval}
\label{sec:consensus_ablation}

To isolate the contribution of consensus-based retrieval, we compare CacheFlow when retrieving from only the final transformer layer (\emph{Layer N}) versus using our proposed two-layer consensus between the first and last layers (\emph{Layer N,0}). The consensus signal provides a more stable indicator of saliency across depth, improving retrieval robustness without increasing compute. As shown in Table~\ref{tab:consensus_ablation}, consensus yields \textbf{small but consistent gains} on both QAEgo4D and EgoSchema.

\begin{table}[h!]
\centering
\setlength{\tabcolsep}{8pt}
\caption{\textbf{Ablation of consensus-based retrieval.} Using both first and last layers (N,0) provides a more stable ranking signal than relying on only the final layer.}
\begin{tabular}{lcc}
\toprule
\multirow{2}{*}{Method} &
\multicolumn{1}{c}{QAEgo4D} &
\multicolumn{1}{c}{EgoSchema} \\
\cmidrule(lr){2-2}
\cmidrule(lr){3-3}
& Acc. & Acc. \\
\midrule
CacheFlow (Layer N)   & 50.2 & 27.6 \\
\rowcolor{gray!10}
CacheFlow (Layer N,0) & 50.8 & 27.9 \\
\bottomrule
\end{tabular}
\label{tab:consensus_ablation}
\vspace{-0.4em}
\end{table}

\subsection{Qualitative Analysis}
\label{sec:qualitative}

\begin{figure}[h!]
\centering
\includegraphics[width=\columnwidth]{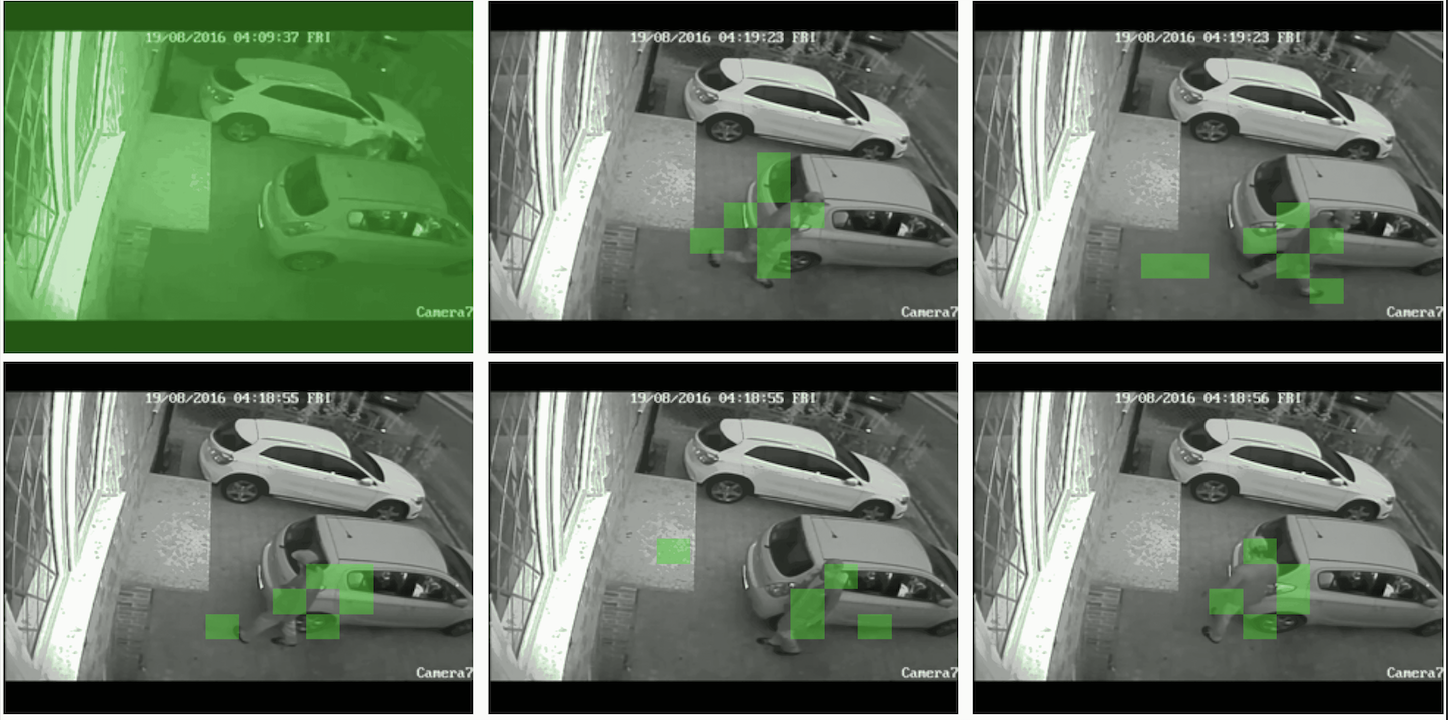}
\caption{\textbf{Qualitative example from MLVU.} Visualizes how
Dynamic Token Dropping (DTD) preserves only the salient regions (green overlays) corresponding to the suspect interacting with the vehicle. Note the first frame is fully preserved by default.}
\label{fig:qualitative}
\end{figure}

To better understand how CacheFlow maintains temporal coherence despite extreme compression, we give an example of its token-selection on MLVU. Figure~\ref{fig:qualitative} shows an example from the surveillance anomaly-detection video in MLVU. The clip contains an irregular event (a person attempting to steal a vehicle), and the question asks: \emph{``Are there any irregularities in this surveillance video? If so, what sort are they?"} with the correct answer being \emph{Stealing}.

CacheFlow identifies the correct answer while discarding \textbf{99.2\%} of all visual tokens (reducing 37{,}828 original tokens to only 305 retained tokens) demonstrating the model's ability to reason over long-range events under extreme compression. The six frames shown in Figure~\ref{fig:qualitative} illustrate how Dynamic Token Dropping (DTD) highlights only the semantically critical regions (green patches), corresponding precisely to the suspect interacting with the vehicle. Despite operating on as few as \textbf{1--7 tokens} in most of the key evidence frames, CacheFlow correctly localizes and preserves the action cues necessary to classify the irregularity.

Notably, the vanilla LLaVA-OV-0.5B baseline (without CacheFlow) answers this sample incorrectly, predicting \textit{``Arrest."} Thus, the qualitative evidence mirrors the quantitative improvements observed in Tables~\ref{tab:cross_benchmark_results} and~\ref{tab:rvs_streaming_results}, where CacheFlow \textbf{delivers $+2$--$7$ point gains} on long-horizon benchmarks while \textbf{discarding 70--87\% of tokens} in streaming settings.

Overall, these qualitative findings support a central claim of this work: CacheFlow retains \emph{event-level narrative continuity} even when operating with a drastically reduced set of visual tokens. The combination of DTD's spatial selectivity and GRU-based temporal summarization enables retrieval of the precise frames needed for causal reasoning, allowing the model to solve long-range video questions that sliding-window or dense-attention baselines fail to address.

\section{Conclusion}
\label{sec:conclusion}
We present \textbf{CacheFlow}, a training-free framework for efficient long-video understanding that unifies \textbf{Dynamic Token Dropping (DTD)} and \textbf{GRU-based Compressive Memory (GCM)}.  
Together, these two components form a compact, streaming memory system that retains essential temporal cues while selectively discarding redundant visual tokens.  
Across six diverse long-video QA benchmarks, CacheFlow delivers accuracy on par with or exceeding strong retrieval-augmented baselines while reducing token processing by up to \textbf{87\%} and cutting inference latency by nearly \textbf{2×}.  
By coupling adaptive sparsification with lightweight, order-aware summarization, CacheFlow demonstrates that \textbf{strong temporal reasoning can coexist with extreme efficiency}, making scalable long-form video understanding practical even without extensive hardware. This work can \textit{significantly} lower the barrier to effective long-video analysis.

\paragraph{Limitations and Future Work.}
While \textbf{CacheFlow} substantially advances efficiency and scalability, several open directions remain.  
First, the current DTD threshold \(\tau_{\text{feat}}\) is fixed during inference. Developing an \textit{adaptive}, context-aware variant that adjusts to scene dynamics could further optimize the balance between speed and recall.  
Second, our GRU-based Compressive Memory (GCM) compresses visual keys and values independently of text or audio, leaving room for a unified multi-modal summarizer that reasons jointly across modalities.  Also, CacheFlow currently relies on fixed-length memory blocks—exploring continuous or hierarchical compression schemes may enable even finer temporal abstraction and lifelong retention. We hope this work motivates broader exploration of \textit{compressive, retrieval-aware architectures} that make long-form video understanding both efficient and enduring.


{
    \small
    \bibliographystyle{ieeenat_fullname}
    \bibliography{main}
}


\end{document}